\pdfoutput=1

\documentclass[11pt]{article}

\usepackage[]{EMNLP2022}
\usepackage{times}
\usepackage{latexsym}

\usepackage[T1]{fontenc}

\usepackage[utf8]{inputenc}

\usepackage{microtype}

\usepackage{inconsolata}

\usepackage{subfigure}
\usepackage{graphicx}
\usepackage{subfigure}
\usepackage{multirow}
\usepackage{makecell}
\usepackage{booktabs}
\usepackage{amsmath}
\usepackage{bm}
\usepackage{etoolbox}
\title{Metric-guided Distillation: Distilling Knowledge from the Metric to Ranker and Retriever for Generative
Commonsense Reasoning }

\author{%
  Xingwei He\textsuperscript{\rm{1}}\thanks{\ \ Work done during internship at Microsoft Research Asia.}, \quad
  Yeyun Gong\textsuperscript{\rm 2}\thanks{\ \ Corresponding author.}, \quad
  A-Long Jin\textsuperscript{\rm 1}, \quad
  Weizhen Qi\textsuperscript{\rm 3}, \quad
  Hang Zhang\textsuperscript{\rm 2}, 
  \\
  \textbf{Jian Jiao,}\textsuperscript{\rm 4} \quad 
  \textbf{Bartuer Zhou,}\textsuperscript{\rm 2} \quad  
  \textbf{Biao Cheng,}\textsuperscript{\rm 2} 
  \quad \textbf{Siu Ming Yiu,}\textsuperscript{\rm 1} 
  \quad \textbf{Nan Duan}\textsuperscript{\rm 2}
  \\
  \textsuperscript{\rm 1}The University of Hong Kong,
  \textsuperscript{\rm 2}Microsoft Research Asia,\\
  \textsuperscript{\rm 3}University of Science and Technology of
China,
  \textsuperscript{\rm 4}Microsoft 
  \\
  \texttt{hexingwei15@gmail.com},  
  \texttt{ajin@eee.hku.hk}, \\
  \texttt{smyiu@cs.hku.hk}, 
  \texttt{weizhen@mail.ustc.edu.cn},  \\
  \texttt{\{yegong, v-zhhang, jian.jiao, bazhou, bicheng, nanduan\}@microsoft.com} 
}

\begin{document}
\maketitle
\begin{abstract}
Commonsense generation aims to generate a  realistic sentence describing 
a daily scene under the given concepts, which is very challenging, since it requires models to have  relational reasoning and compositional generalization capabilities.
Previous work focuses on retrieving prototype sentences for the provided concepts to assist generation. 
They first use a sparse retriever to retrieve candidate sentences, then re-rank the candidates with a ranker. 
However, the candidates returned by their ranker may not be the most relevant sentences, 
since the ranker treats all candidates equally without considering their relevance to the reference sentences of the given concepts. Another problem is that re-ranking is very expensive, 
but only using retrievers will seriously degrade the performance of their generation models. 
To solve these problems, we propose the metric distillation rule to distill knowledge from the metric (e.g., BLEU) to the ranker. 
We further transfer the critical knowledge summarized by the distilled ranker to the retriever. 
In this way, the relevance scores of candidate sentences predicted by the ranker and retriever will be more consistent with their quality measured by the metric. 
Experimental results on the CommonGen benchmark verify the effectiveness of our proposed method: (1) Our generation model with the distilled ranker achieves a new state-of-the-art result. (2) Our generation model with the distilled retriever even surpasses the previous SOTA.  

\end{abstract}

\section{Introduction}
Commonsense reasoning is the ability to make reasonable and logical assumptions about daily scenes, which is a long-standing challenge in natural language processing. 
Recently, many discriminative tasks,  such as CommonsenseQA \cite{talmor-etal-2019-commonsenseqa} and SWAG \cite{sap-etal-2019-social}, have been proposed to evaluate the commonsense reasoning ability by testing whether models can select the correct answer from the choices according to the given context.  
To test whether models acquire the generative commonsense reasoning ability, \citet{commongen} proposed the commonsense generation (CommonGen) task, which requires models to produce a plausible sentence describing a specific daily life scenario based on the given concepts. 

CommonGen proposes two main challenges to models, and it expects models to (1) reason over the commonsense relations among concepts to generate sentences in line with our commonsense; 
(2) possess the compositional generalization ability to generate realistic sentences with unseen concept compositions. 
Experiment results \cite{commongen} show that large-scale pre-trained models (e.g., BART) alone is not competent for this task (see Table \ref{tab:example}).  
The main reason is that the source information is very limited; therefore, the models can only rely on the internal implicit knowledge acquired during pre-training to solve this problem, resulting in generating some sentences that violate commonsense.

\begin{table}
  \scriptsize
    \centering
      \begin{tabular}
      {
       m{0.46\textwidth}<{\raggedright}
       }
      \toprule

     \textbf{Concepts}:  {\textbf{eye}, \textbf{hang}, \textbf{head}, \textbf{shut}, \textbf{squeeze}}  \\
      \textbf{Reference}: A man \textbf{squeezes} his \textbf{eyes} \textbf{shut} and \textbf{hangs} his \textbf{head}.  \\
      \midrule
      \textbf{BART}: He \textbf{squeezes} her \textbf{head} \textbf{shut}, then grasps her \textbf{eyes} shut.      \\
      \textbf{Our}: A baby with a blue shirt \textbf{hangs} his \textbf{head} and \textbf{squeezes} his \textbf{eyes} \textbf{shut}.   \\
    
      \bottomrule

    \end{tabular}
    \caption{ Sentences generated by BART and our proposed model, DKMR$^2$.
    }\label{tab:example}
\end{table}

To enrich the source information, EKI-BART \citep{fan2020enhanced} first retrieves prototype sentences for the input concepts, 
and then feeds the concepts and retrieved sentences into the generation model.  
Recent work, such as RE-T5 \cite{wang-etal-2021-retrieval-enhanced}, KFCNet \cite{li-etal-2021-kfcnet-knowledge}, and KGR$^4$ \cite{liu2021kgr}, 
extends this retrieve-and-generate framework by introducing a binary classifier to re-rank the retrieved candidate sentences and filter out candidates irrelevant to the input concepts. 
One problem with these works is the discrepancy between training and re-ranking for their ranker. 
Concretely, when training the ranker, they treat all retrieved candidate sentences as negatives, regardless of their relevance to the reference sentences of input concepts. 
However, during re-ranking, the ranker is asked to point out how these candidates differ in their relevance to references. 
Another problem is that the re-ranking process of the cross-encoder ranker is very time-consuming, which is non-negligible, especially for online systems. 

In this paper, we also resort to the retrieve-and-generate pipeline to solve CommonGen, yet further improve the retrieval module by alleviating the above problems.
Our motivation is to expect that the relevance scores of candidates computed by the ranker and retriever are in line with the gold quality scores between candidates and reference sentences measured with the evaluation metric. 
To achieve this, we first distill the gold rank knowledge of candidates measured by the  metric to the ranker. 
Next, we improve the retriever by transferring the metric knowledge from the distilled ranker to the retriever 
rather than directly distilling it from the metric (please refer to Section \ref{distill_retriever} for more explanation). 
By doing so, the distilled ranker and retriever can select more relevant sentences than  their counterparts without metric distillation.  

The contributions of this work are summarized as follows: 
(1) We propose to \textbf{D}istill \textbf{K}nowledge from the \textbf{M}etric to \textbf{R}anker and \textbf{R}etriever, termed DKMR$^2$, for generative commonsense reasoning, 
which uses the metric-guided distillation to improve the ranker and a progressive distillation strategy to improve the retriever\footnote{Our code and models are available at  https://github.com/microsoft/advNLG.}.  
(2) We conduct extensive experiments on the CommonGen benchmark. Our proposed model 
achieves a new state-of-the-art (SOTA) on both the v1.0 test set (43.37 vs. 39.15 on SPICE) and the official test set (v1.1) (34.589 vs. 33.911 on SPICE) of the leaderboard.
(3) 
The performance of DKMR$^2$ with the distilled retriever is on par with DKMR$^2$ using the distilled ranker. 
As a result, the expensive retrieve-then-rank pipeline can be replaced with the distilled retriever at the expense of negligible performance. 
 
\section{Problem Statement}
CommonGen is a constrained text generation task, with the goal of generating a coherent and plausible sentence $\bm{s}$ describing an everyday scenario using an unordered concept set $\bm{c}=\{c_1, c_2 \dots, c_m\}$. Therefore, this task is typically formulated to maximize the conditional probability of $\bm{s}$:  
\begin{eqnarray}\label{eq:1}
  p(\bm{s}|\bm{c}; \theta) = \prod_{t=1}^{n}p(s_t|s_{i<t}, \bm{c};\theta),
\end{eqnarray} 
where $n$ denotes the length of the generated sequence $\bm{s}$ and $s_{i<t}$ refers to the sub-sequence generated before the time step $t$.

\section{Methodology}
\begin{figure*}
  \centering

    \includegraphics[width=1\textwidth]{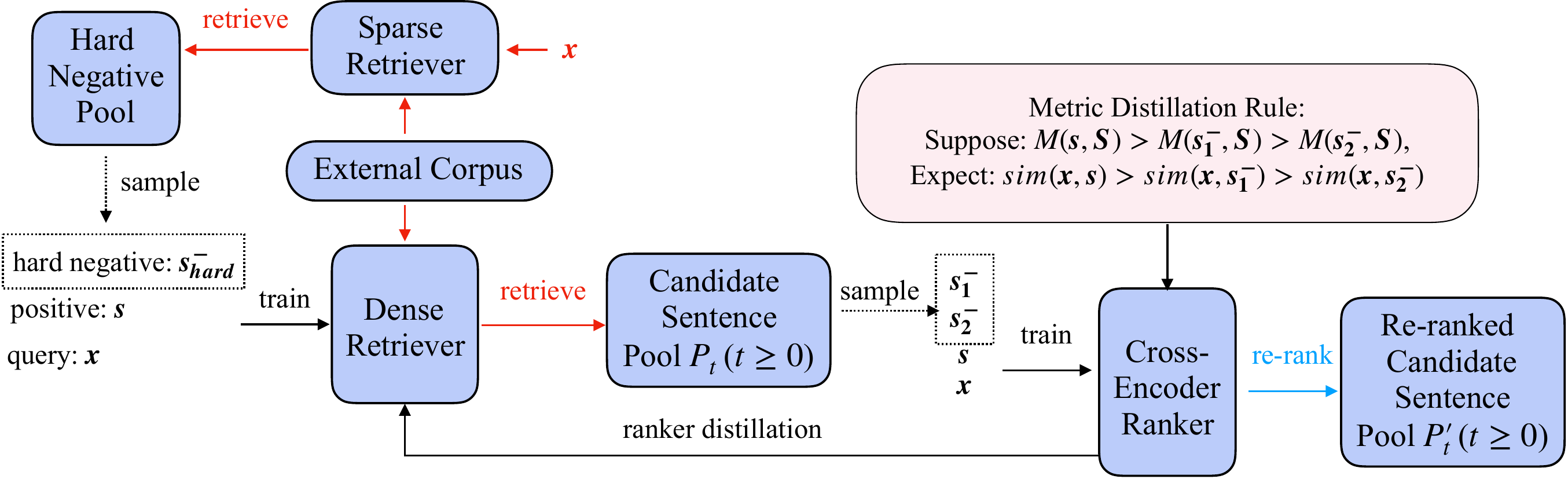} 
      \caption{ 
            The pipeline of the retrieval module. Dotted, red, black and blue lines denote the `sample', `retrieve', `train' and `re-rank' processes, respectively. $\bm{x}$ and $\bm{s}$ are one paired source and target from CommonGen.  
             $\bm{S}$ denotes the reference sentences ($\bm{s}\in \bm{S}$). $M$ is the automatic evaluation metric, used to measure the quality of the retrieved sentence in terms of $\bm{S}$. 
      }
      \label{pipeline}
\end{figure*}

Following the previous work \cite{fan2020enhanced}, we resort to the retrieve-then-generate framework to solve CommonGen, which mainly consists of two modules, the retrieval module and the generation module. 
The retrieval module aims to retrieve relevant sentences to assist the generation module in generating desirable outputs. 
Recent work \cite{wang-etal-2021-retrieval-enhanced, li-etal-2021-kfcnet-knowledge, liu2021kgr} extends this idea by introducing a ranker to the retrieval module. 
In this work, our retrieval module also resorts to the retrieve-then-rank pipeline, as illustrated in Figure \ref{pipeline}. To be specific, the retrieval module mainly contains two models, the retriever  and ranker, where the retriever is used to retrieve candidate sentences for the given concept set and the ranker further re-ranks the retrieved sentences. Different from previous work, we improve the ranker by distilling knowledge from the gold quality scores between the candidate sentences and gold reference sentences computed by the evaluation metric. Then, the distilled ranker will pass the distilled knowledge to the retriever, aiming to correct the retriever's inaccurate retrieval operations. 

In Section \ref{retriever}, we first introduce the warm-up of the retriever. Then, we will show how to distill knowledge from the metric, in turn, to the ranker and retriever in Sections \ref{distill_ranker} and \ref{distill_retriever}, respectively. 
Finally, we will show how to generate sentences based on the retrieved sentences in Section \ref{generator}.

\subsection{Warm-up of the Retriever}\label{retriever} 
We use a typical dense retrieval model \cite{karpukhin-etal-2020-dense} as the retriever. As shown in Figure \ref{retriever_ranker}(a) in  Appendix \ref{framework}, the retriever is based on the dual-encoder architecture.  
In this work, we implement the retriever with two independent encoders, initialized with BERT. We use the hidden state of the first token (i.e., [CLS]) at the last layer as the representation of the input sentence.
We first use the sentence encoder $E_s(.)$ to compute the $d$-dimensional dense representations for all sentences in the external corpus $D$. Then, we use the concept encoder $E_c(.)$ to compute the dense representation of the concept set. The similarity $sim(\bm{c}, \bm{s})$ between them is measured by the dot product of their dense representation vectors:
\begin{eqnarray}
  sim(\bm{c}, \bm{s}) = E_c(\bm{c})^T E_s(\bm{s}).
\end{eqnarray} \\
\textbf{Train.} During training, we warm up the dual-encoder retriever, Retriever$_0$, with the contrastive loss \cite{pmlr-v119-chen20j}:
\begin{equation}
\begin{aligned}
  &L(\bm{c}; \bm{s}, \bm{s_1}, \dots, \bm{s_N^-})  \nonumber \\ 
  &=-log\frac{e^{(sim(\bm{c}, \bm{s}))}}{exp^{(sim(\bm{c}, \bm{s}))} + \sum_{i=1}^{N} exp^{(sim(\bm{c}, \bm{s_i^-})) }},\\
\end{aligned}
\end{equation} 
where $sim(\bm{c}; \bm{s})$ denotes the relevance score between the concept set $\bm{c}$ and the positive sentence $\bm{s}$. Similarly, $sim(\bm{c}; \bm{s_i^-})$ is the relevance score between $\bm{c}$ and $i-$th negative sentence. $N$ refers to the number of negative sentences. 

Following DPR \cite{karpukhin-etal-2020-dense}, the negative sentences consist of one hard negative  and $N-1$ in-batch negatives\footnote{Note that in-batch negatives come from other positive target sentences in a mini-batch. Therefore, $N$ equals the batch size in one GPU card.}. We consider two sparse retrievers to build the hard negative pool: (1) TF-IDF: compute the similarity scores between the sparse vectors of $\bm{c}$ and sentences in $D$; (2) Concept matching: sort sentences in $D$ according to the number of concepts appearing in each sentence in descending order. Each sparse retriever will return the top $K$ sentences as hard negative pool $P$ for one concept set. When training Retriever$_0$, we randomly sample one from $P$ as the hard negative. 
\\
\textbf{Retrieve. }
During the retrieval stage, we first compute the sentence representations for all sentences in $D$ with the sentence encoder $E_s(.)$. To accelerate the retrieval process, we build IndexFlatIP indexes for representation vectors with the FAISS \cite{johnson2019billion} library, which is efficient for approximate nearest neighbor similarity search for billions of dense vectors. We return the top $K$ sentences for each concept set with Retriever$_0$ as the candidate sentence pool $P_0$. 
We find that when training the retriever with hard negatives from concept matching, 
$P_0$ is more helpful to the generation model (refer to Appendix \ref{hardneg} for more details).

\subsection{Distilling Knowledge from the Metric to the Ranker} \label{distill_ranker}
Our ranker is based on the cross-encoder architecture, as shown in Figure \ref{retriever_ranker}(b) in Appendix \ref{framework}. We implement the ranker with BERT by putting a forward layer over the hidden state of [CLS] at the last layer. The one-dimensional output is regarded as the similarity score between the concept set and the candidate sentence, $sim(\bm{c}, \bm{s})$. 

Previous work uses the binary cross-entropy loss or contrastive loss to train rankers. 
During training, these works treat all negatives equally, without distinguishing the differences between them, but during the re-ranking period, they expect rankers to tell the differences between candidate sentences. \\
\textbf{Metric Distillation Rule.}
To bridge the gap between training and re-ranking, we propose the metric distillation rule to distill knowledge from the metric to the ranker. Suppose $\bm{s}$ is one positive sentence, $\bm{S}$ is the reference sentences, $\bm{s_1^-}$ and $\bm{s_2^-}$ are two negatives for the given concept set $\bm{c}$. 
Based on an automatic evaluation metric $M$ (e.g., BLEU or ROUGE), the gold quality scores of sentences are ranked like this: 
$M(\boldsymbol{s}, \boldsymbol{S})>M(\boldsymbol{s_1^-}, \boldsymbol{S})>M(\boldsymbol{s^-_2}, \boldsymbol{S})$. 
We expect the order of relevance scores measured by the ranker to be consistent with the quality order measured by the metric $M$:   
$sim(\boldsymbol{x}, \boldsymbol{s})>sim(\boldsymbol{x},\boldsymbol{s^-_1})>sim(\boldsymbol{x}, \boldsymbol{s^-_2})$,
which is defined as the metric distillation rule\footnote{We empirically find that instructing the ranker to learn the order knowledge is better than learning exact quality scores. }. \\
\textbf{Train.} During training, the ranker is guided by the metric distillation rule. Therefore,  the quality scores of sentences measured by $M$ serve as the teacher, while the ranker is a student. To fulfill the metric distillation rule, we resort to the ListMLE loss \cite{xia2008listwise} to optimize the ranker: 
\begin{eqnarray}
\begin{aligned}
&\bm{z} = [sim(\bm{c},\bm{s_1}), \dots, sim(\bm{c},\bm{s_{N_1}})] \\
&L_{ListMLE} =-log\prod_{k=1}^{N_1}\frac{e^{(z_{o_k})}}{\sum_{i=k}^{N_1}e^{(z_{o_i})  }}, \\
\end{aligned}\label{eq3}
\end{eqnarray}
where $o_i$ is the quality order of the $i$-th sentence measured by $M$. $N_1$ is the number of sentences. \\
\textbf{Re-rank.} At the $t$-th ($t\geq 0$) re-ranking stage, we sort sentences in the candidate sentence pool $P_t$ with the ranker Ranker$_t$ and output them into the re-ranked candidate sentence pool $P_t^{\prime}$.

\subsection{Distilling Knowledge from the Ranker to the Retriever} \label{distill_retriever} 
Based on the above discussion, we can readily find that the warm-up retriever Retriever$_0$ also suffers from the discrepancy between training and retrieving. Intuitively, we can mitigate this problem by directly distilling knowledge from the metric to the retriever in a similar way to the ranker, yet we find that it is much better to distill knowledge from the distilled ranker. In other words, the \textbf{progressive distillation} path (i.e., `metric->ranker->retriever') is superior to the direct distilling path (i.e., `metric->retriever'). One possible explanation for this counter-intuitive phenomenon is that  the quality distribution of candidate sentences measured by the metric is complex, but the retriever's learning ability is limited. 
In contrast, the ranker has a stronger data fitting ability. Compared with the gold quality distribution, the output of the distilled ranker is much smoother, making it easier for the retriever to acquire. 
For ease of understanding, we make the following analogy:  a knowledgeable person (e.g., a university professor or an expert in some field) may not be a suitable teacher for a novice (e.g., a primary school student or a beginner in the field). 
In this case, it may be much better to find an intermediary with strong learning ability. The intermediary first learns from the knowledgeable person and then teaches the novice the simplified knowledge points. 
\\
\textbf{Train.} During training, we distill knowledge from the ranker to the retriever by minimizing the Kullback–Leibler (KL) divergence:
\begin{eqnarray}
\begin{aligned}
\bm{z} &= [sim(\bm{c},\bm{s_1}), \dots, sim(\bm{c},\bm{s_{N_2}})]  \\ 
\bm{l} &= [sim^\prime(\bm{c},\bm{s_1}), \dots,  sim^\prime(\bm{c},\bm{s_{N_2}})]   \\
L_{KL} &= KL(\bm{z}|| \bm{l}), \\
\end{aligned}\label{eq4}
\end{eqnarray}
where $sim$ and $sim^\prime$ denote the similarity scores between the concept set and candidate sentence computed by the ranker and retriever , respectively. $N_2$ is the number of sentences. \\
\textbf{Retrieve.} The retrieval stage is the same with that in Section \ref{retriever}. 
\subsection{Retrieval-Augment Text Generation} \label{generator}
\begin{figure}
  \centering
      \includegraphics[width=0.48\textwidth]{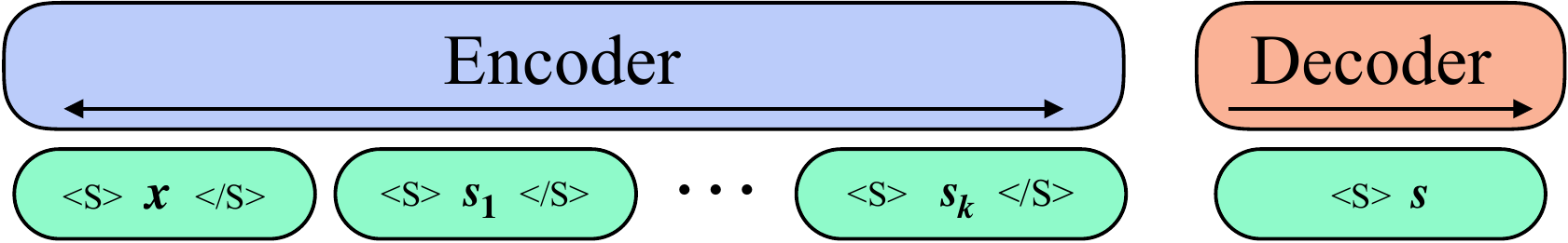} 
      \caption{ 
            The overview of the generator. <S> and </S> are special tokens denoting the start and end of a sentence, respectively. $\bm{x}$ denotes the concept set. $\bm{s_1}, \dots, \bm{s_k}$ are the retrieved sentences. $\bm{s}$ is the target sentence from CommonGen.
      }
      \label{fig:generator}
\end{figure}

During generation, we select the top $k$ sentences from a candidate sentence pool to help the generator generate target sentences. We concatenate the concept set and the retrieved sentences, and directly feed them into the encoder of the generator (see Figure \ref{fig:generator} for the input format). During training, we optimize the generator by minimizing the cross-entropy loss between the predicted sentence of the decoder and the golden sentence.

\section{Experiments}

\subsection{Experimental Setups}
\textbf{Dataset.}
Following previous work, we conduct experiments on the CommonGen dataset released by \citet{commongen}.
Table \ref{tab:data} shows the basic statistics of this dataset. As shown in Table \ref{tab:data}, most concept compositions of the validation and test sets are unseen in training data, posing a compositional generalization challenge to models. 

As mentioned above, our model also needs an external corpus. 
To make a fair comparison with the previous work \citep{li-etal-2021-kfcnet-knowledge}, 
we construct the external corpus from the same sources: ActivityNet \citep{krishna2017dense},
VaTeX \citep{wang2019vatex}, Conceptual Captions \citep{sharma-etal-2018-conceptual}, SNLI \citep{bowman-etal-2015-large}, and MNLI \citep{williams-etal-2018-broad}. 
We filter out the sentences containing more than 20 or less than four words. We also remove the sentences that appear in the CommonGen dataset. 
After filtering, about 3.8M sentences are left, which are used as the external corpus. 
\begin{table}
   \footnotesize
    \centering
      \begin{tabular}
      {
       m{0.2\textwidth}<{\raggedright}|
       m{0.055\textwidth}<{\raggedleft}
       m{0.055\textwidth}<{\raggedleft}
       m{0.055\textwidth}<{\raggedleft}
       }
      \toprule

      Partition & Train & Validation & Test \\
      \midrule

      \#Concept Sets& 32,651 & 993 & 1,497\\   
      \#Sentences & 67,389 & 4,018& 6,042 \\ 
      Avg. Sentence Length &10.54 & 11.55 &13.34\\ 
      \midrule
      Unseen Concepts &- &6.53\% & 8.97\% \\
     Unseen Concept-Pairs &- &96.31\% & 100.00\% \\
     Unseen Concept-Triples &- &99.60\% & 100.00\% \\

    \bottomrule
    \end{tabular}
    \caption{ The basic statistics of the CommonGen dataset\footnotemark.
    \#Concept Sets and \#Sentences denote the number of unique concept sets and sentences, respectively.  The unseen compositions refer to the ratios of unique concept, concept-pair, and concept-triple without appearing in the training set. High percentages of unseen concept compositions enable this task to validate the generalization ability of different models effectively. 
    }\label{tab:data}
\end{table}
\footnotetext{Note that \# Sentences denotes the number of sentences of the v1.0 test set. Shortly after, \citet{commongen} released the second version of the test set (v1.1). Compared with the v1.0 test set, the v1.1 test set has one more human reference for each concept set (previously 4, now 5), yet the concept sets for all data sets are unchanged. Since the v1.1 test set is not publicly available, we mainly show the experimental results on the v1.0 test set.
}
\\
\textbf{Evaluation Metrics.} 
Following \citet{commongen}, we resort to BLUE \citep{Papineni2002BleuAM}, ROUGE \citep{lin-2004-rouge}, and METEOR \citep{banerjee-lavie-2005-meteor} to measure the surface similarities between the generated sentences and human references. In addition, we use CIDEr \citep{Vedantam_2015_CVPR}
and SPICE \citep{anderson2016spice} to evaluate the generated sentences.  By comparison, these two metrics aim to assess the correlations among mentioned concepts rather than n-gram overlaps. 
Following \citet{li-etal-2021-kfcnet-knowledge}, we also compute the average score across all metrics as the overall score\footnote{The official evaluation code for CommonGen is available at  https://github.com/INK-USC/CommonGen.}. \\
\textbf{Baselines.} We compare our proposed model with two kinds of strong
baselines. The first kind of baselines is based on large pre-trained models, including GPT-2, BERT-Gen, UniLM, BART,
and T5, implemented by \citet{commongen}. They directly fed the concatenated concepts (e.g. ``$c_1\ \ c_2\ \ \dots\ \ c_k$'')
as input to the first four pre-trained models. As for T5, they prepended the concatenated concepts with a prompt and fine-tuned T5-based models on the format ``generate a sentence with $c_1\ \ c_2\ \ \dots\ \ c_k$.''   

Another kind of baseline uses external knowledge to boost the performance further. 
KG-BART \citep{liu2021kg} incorporates a knowledge graph into both the encoder and decoder, which provides rich relational information among the concepts. 
Different from KG-BART, EKI-BART, RE-T5, KFCNet, and KGR$^4$ apply the retrieve-and-generate framework for CommonGen. \\
\textbf{Implementation Details.} Our implementation details are shown in Appendix \ref{imp}.

\begin{table*}[t] 
  \centering
 \footnotesize
   \begin{tabular}{
    m{0.3\textwidth}<{\raggedright}
    m{0.05\textwidth}<{\centering}
    m{0.05\textwidth}<{\centering}
    m{0.05\textwidth}<{\centering}
    m{0.05\textwidth}<{\centering}
    m{0.07\textwidth}<{\centering}
    m{0.07\textwidth}<{\centering}
    m{0.06\textwidth}<{\centering}
    m{0.06\textwidth}<{\centering}
    }
    \toprule
    \textbf{Models/Metrics} & \multicolumn{2}{c} {\textbf{ROUGE-2/L} } & \multicolumn{2}{c} {\textbf{BLEU-3/4} }& \textbf{METEOR} &\textbf{CIDEr}& \textbf{SPICE}& \textbf{Overall} \\
    \midrule

    GPT-2  \citep{Radford2019LanguageMA} & 17.18 & 39.28 & 30.70 & 21.10 & 26.20 & 12.15 & 25.90 & 24.64 \\
    BERT-Gen  \citep{pmlr-v119-bao20a} & 18.05 & 40.49 & 30.40 & 21.10 & 27.30 & 12.49 & 27.30 & 25.30 \\
    UniLM \citep{NEURIPS2019_c20bb2d9} & 21.48 & 43.87 & 38.30 & 27.70 & 29.70 & 14.85 & 30.20 & 29.44\\
    UniLM-v2  \citep{pmlr-v119-bao20a} & 18.24 & 40.62 & 31.30 & 22.10 & 28.10 & 13.10 & 28.10 & 25.93 \\
    T5-Base  \citep{JMLR:v21:20-074} & 14.57 & 34.55 & 26.00 & 16.40 & 23.00 & 9.16 & 22.00 & 20.81  \\
    T5-Large  \citep{JMLR:v21:20-074} & 22.01 & 42.97 & 39.00 & 28.60 & 30.10 & 14.96 & 31.60 & 29.89 \\
    BART-Large  \citep{Lewis2020BARTDS} & 22.23 & 41.98 & 36.30 & 26.30 & 30.90 & 13.92 & 30.60 & 28.89 \\
    
    \midrule
    KG-BART \citep{liu2021kg} & 23.38 & 44.54 & 42.10 & 30.90 & 32.40 & 16.83 & 32.70 & 31.83 \\
    EKI-BART \citep{fan2020enhanced} & 25.43 & 46.53 & 46.00 & 36.10 & 33.80 & 17.80 & 33.40 & 34.15 \\
    KFCNet \citep{li-etal-2021-kfcnet-knowledge} & 26.81 & 47.52 & 57.33 & 51.46 & 38.92 & 20.98 & 39.15 & 40.31 \\
    
    \midrule
    DKMR$^2$ + Retriever$_1$ &28.64 & 48.61 & 68.00& 62.50& 44.59& 24.25 & 42.42 & 45.57\\
        
    DKMR$^2$ + Ranker$_{0}$& \textbf{29.33} & \textbf{49.22}& \textbf{69.48}& \textbf{64.19}& \textbf{46.01}& \textbf{24.85}& \textbf{43.37} & \textbf{46.64}\\

    \bottomrule
 \end{tabular}
 \caption{Results of different models on the CommonGen test set (v1.0). 
 The top of the table shows the results of different pre-trained models (the result numbers are extracted from \citet{commongen}). The middle of the table shows the results of knowledge-enhanced models (results in each row are from the corresponding cited paper). 
 DKMR$^2$+Ranker$_{0}$ means using the sentences returned by the distilled Ranker$_{0}$ to assist generation.  }\label{tab_main_result} 
\end{table*} 

\subsection{Experimental Results}
We show our experimental results on the CommonGen test set in Table \ref{tab_main_result}, from which we can draw the following conclusions: \\
(1) \textbf{Only using large-scale pre-trained models is not sufficient to solve this task.} From the top of Table \ref{tab_main_result}, we observe that simply feeding the concepts into pre-trained models does not produce satisfactory results. This shows that pre-trained models cannot infer the relationship among concepts well only by relying on their internal knowledge. \\
(2) \textbf{Using external knowledge to enrich the source information is an effective strategy for commonsense generation. }
As shown in the middle of Table \ref{tab_main_result}, either using knowledge graphs or retrieved sentences can bring clear performance improvements, 
since they can provide valuable information to help the generation model to reason the relations among concepts.   \\
(3) \textbf{Metric-guided distillation helps instruct the ranker to retrieve more relevant candidate sentences to profit generation. } 
KFCNet is the previous SOTA, which first extracts sentences containing concepts as candidate sentences and then re-ranks them with a binary ranker. For fair comparisons, our proposed model, DKMR$^2$, uses the same external corpus and generation model as KFCNet (we use BART-base, KFCNet uses BART-large). 
The main difference comes from the retrieval module. 
Concretely, DKMR$^2$ considers the quality scores of candidates in terms of the gold reference sentences, which are measured by the automatic evaluation metric, $M(\bm{s},\bm{S})$. 
We train the ranker by distilling the rank knowledge from the quality scores measured by the metric. 
As shown in Table \ref{tab_main_result}, our proposed model, DKMR$^2$+Ranker$_0$, outperforms KFCNet on all metrics by a large margin and achieves a new state-of-the-art result, which proves the effectiveness of the metric distillation strategy.

\begin{table}[t] 
  \centering
 \footnotesize
   \begin{tabular}{
    m{0.09\textwidth}<{\raggedright}
    m{0.07\textwidth}<{\raggedright}|
    m{0.08\textwidth}<{\centering}
    m{0.05\textwidth}<{\centering}
    m{0.05\textwidth}<{\centering}
    }
    \toprule
    \textbf{Variants} &\textbf{Distill} & \textbf{BLEU-4} & \textbf{CIDEr} & \textbf{SPICE} \\
    \midrule

    Retriever & & 55.50 & 21.96 & 39.93 \\
    Retriever & $\surd$&  62.50 & 24.25& 42.42 \\
    \midrule
    Ranker &  & 60.27 & 23.30& 42.26 \\
    Ranker & $\surd$ & \textbf{64.19} & \textbf{24.85} & \textbf{43.37}  \\
    \bottomrule
 \end{tabular}
 \caption{
 Results of our generation model with candidates returned by retriever or ranker with/without distillation on the CommonGen test set (v1.0). 
 }\label{tab_result2} 
\end{table}

\subsection{Ablation Study and Analysis}
In this section, we mainly test the effect of different retrieval methods. We use different strategies to train retrievers or rankers and then use the sentences returned by them to help the BART-based generation model produce target sentences. \\
\textbf{Effect of the Metric Distillation.} 
To clearly demonstrate the impact of the metric distillation on the retriever and ranker, we show two groups of comparative experiments in Table \ref{tab_result2}, from which we can see that the metric-guided distillation brings improvements to both the ranker and retriever, especially to the retriever. For example,  the distilled retriever (i.e., Retriever$_1$) increases SPICE by around 2.5 points in contrast to its counterpart without distillation (i.e., Retriever$_0$). 

More importantly, the metric distillation narrows the performance gap between the ranker and retriever. The distilled retriever is even on a par with the distilled ranker (e.g., 42.42 vs. 43.37 on SPICE). 
This brings us a significant advantage: when we deploy the online system, we can replace the time-consuming retrieve-then-rank pipeline with the distilled retriever with only negligible performance loss. 
In addition, we find that one distillation step is enough for the ranker and retriever, and we do not observe any significant improvements in the continuous distillation steps.  
\begin{table}[t] 
  \centering
 \footnotesize
   \begin{tabular}{
    m{0.18\textwidth}<{\raggedright}|
    m{0.08\textwidth}<{\centering}
    m{0.05\textwidth}<{\centering}
    m{0.05\textwidth}<{\centering}
    }
    \toprule
    \textbf{Ranker$_0$ Variants}&\textbf{BLEU-4} & \textbf{CIDEr} & \textbf{SPICE} \\
    \midrule
    w/o Distillation & 60.27 & 23.30& 42.26 \\
    \midrule
    w/ KL & 62.40 & 24.06 & 42.39\\
    w/ ListMLE & \textbf{64.19}& \textbf{24.85}& \textbf{43.37} \\
    \bottomrule
 \end{tabular}
 \caption{Results of training Ranker$_0$ with different distillation strategies on the CommonGen test set (v1.0). KL denotes distilling all knowledge with the KL-divergence loss. 
 ListMLE denotes distilling the order knowledge with the ListMLE loss. 
 }\label{tab_result3} 
\end{table} 
\\
\textbf{Effect of Distilling the Order Knowledge.} 
We conduct experiments to compare the effect of different distillation strategies used to distill knowledge from the metric to the ranker. 
As shown in Table \ref{tab_result3}, both distilling strategies outperform their counterpart without distilling knowledge from the metric. 
In addition, the ranker trained by distilling only the order knowledge from the metric even performs better than its counterpart distilling all knowledge. 
We speculate that this may be because the quality distribution of candidates evaluated by the metric is too complex for the ranker to learn accurately. 
In fact, learning only the order relation does not affect the model to pick out the best sentence from  candidates in theory, yet dramatically reduces the learning difficulty. 
\\
\textbf{Effect of the Progressive Distillation.} 
To demonstrate the effect of the progressive distillation strategy for the retriever, we train the retriever Retriever$_1$ by (1) directly distilling knowledge from the metric or (2) distilling from the distilled ranker. 
As shown in Table \ref{tab_result4}, both distilled retrievers far exceed the retriever without using distillation. 
In addition, we also find that the progressive distillation strategy performs much better than the direct distillation strategy. 

As stated in Section \ref{distill_retriever}, the dual-encoder retriever has a limited learning ability since it cannot capture fine-grained interactions between the concepts and sentences. 
On the other hand, the quality distribution of candidate sentences measured by the metric may be too complex. 
In light of these facts, directly learning from the complex distribution may overwhelm the retriever.
Compared with the retriever, the cross-encoder ranker has a much stronger learning ability. 
Therefore, the ranker is more qualified to learn from the metric and then transfer the learned knowledge to the retriever.  
In this way, the retriever does not need to learn all tedious details from the metric, and only needs to acquire the critical points summarized by the ranker, which undoubtedly relieves the retriever's learning burden pressure. 
\begin{table}[t] 
  \centering
 \footnotesize
   \begin{tabular}{
    m{0.19\textwidth}<{\raggedright}|
    m{0.08\textwidth}<{\centering}
    m{0.05\textwidth}<{\centering}
    m{0.05\textwidth}<{\centering}
    }
    \toprule
    \textbf{Retriever$_1$ Variants} &\textbf{BLEU-4} & \textbf{CIDEr} & \textbf{SPICE} \\
    \midrule
    w/o Distillation & 55.50 & 21.96 & 39.93 \\
    \midrule
    Distilling from Metric  & 58.54&  22.76& 41.04 \\
    Distilling from Ranker$_0$ & \textbf{62.50} & \textbf{24.25}& \textbf{42.42} \\
    \bottomrule
 \end{tabular}
 \caption{ Results of training Retriever$_1$ with different distillation sources on the CommonGen test set (v1.0). 
 }\label{tab_result4} 
\end{table} 
\begin{table}[t] 
  \centering
 \footnotesize
   \begin{tabular}{
    m{0.09\textwidth}<{\raggedright}|
    m{0.08\textwidth}<{\centering}
    m{0.08\textwidth}<{\centering}
    m{0.05\textwidth}<{\centering}
    m{0.05\textwidth}<{\centering}
    }
    \toprule
    \textbf{DM}  &\textbf{BLEU-4} &\textbf{METEOR} & \textbf{CIDEr} & \textbf{SPICE} \\
    \midrule
    -  & 60.27 & 44.42 & 23.30 & 42.26 \\
    BLEU  & 64.19  & 46.01 & 24.85  & \textbf{43.37} \\
    METERO & 63.00 & 45.15 & 24.36 & 42.64\\
    ROUGE-2 & \textbf{65.15} &\textbf{46.29} & \textbf{24.88} & 43.18 \\
    ROUGE-L & 65.03 &45.97 & 24.85 & 43.24 \\
    \bottomrule
 \end{tabular}
 \caption{Results of training Ranker$_0$ with different distillation metrics (DM) on the test set (v1.0). `-' denotes without using distillation metrics. 
 }\label{tab_result5} 
\end{table} 
\\
\textbf{Effect of Distillation Metrics.} 
In other experiments, we use BLEU as the distillation metric. 
To test the effect of different distillation metrics, we train the ranker with other metrics. 
As shown in Table \ref{tab_result5}, all distilled rankers outperform the ranker without distillation (row 1). In addition, rankers distilled with ROUGE-2 and ROUGE-L have a similar performance to the ranker distilled with BLEU. 
These prove that our metric distillation  method is not limited to BLEU, and it can be easily extended to other metrics. \\
\subsection{Official Leaderboard Results}
We also evaluate our proposed model on the official test set (v1.1). Since the latest test set is not publicly available, the results are obtained through official evaluation. 
We select some representative baselines from the official leaderboard\footnote{https://inklab.usc.edu/CommonGen/leaderboard.html} 
and show their results in Table \ref{tab_result_leaderboard}.  
Consistent with the results on the v1.0 test set, we again observe the followings: 
(1) DKMR$^2$ with a distilled ranker achieves a new  state-of-the-art result. (2) DKMR$^2$ with a distilled retriever even outperforms KFCNet with a ranker. (3) Metric-guided distillation narrows the gap between the ranker and retriever. These observations once again prove the effectiveness of metric distillation.

\begin{table}[t] 
  \centering
 \footnotesize
   \begin{tabular}{
    m{0.18\textwidth}<{\raggedright}
    m{0.08\textwidth}<{\centering}
    m{0.05\textwidth}<{\centering}
    m{0.05\textwidth}<{\centering}
    }
    \toprule
    \textbf{Models/Metrics} & \textbf{BLEU-4} & \textbf{CIDEr} & \textbf{SPICE} \\
    \midrule
    Human (Upper Bound) & 46.49 & 37.64 & 52.43 \\
   \midrule
    DKMR$^2$ + Ranker$_0$  & \textbf{44.334} & \textbf{19.538}& \textbf{34.589} \\
    DKMR$^2$ + Retriever$_1$  & 44.054 &19.353& 34.133 \\
    \hline
    KFCNet & 43.619&	18.845 &	33.911 \\
    KGR$^4$ &42.818&	18.423&	33.564 \\
    RE-T5 & 40.863&	17.663&	31.079 \\ 
    KG-BART & 33.867&	16.927&	29.634 \\
    EKI-BART & 35.945&	16.999&	29.583 \\
    T5-Large & 31.962&	15.128&	28.855 \\
    BART-Large & 31.827	&13.976&	27.995 \\
    UniLM & 30.616&	14.889&	27.429 \\

    \bottomrule
 \end{tabular}
 \caption{Results of different models on the CommonGen test set (v1.1).}\label{tab_result_leaderboard} 
\end{table}

\begin{table*}
  \footnotesize
    \centering
      \begin{tabular}
      {
       m{0.96\textwidth}<{\raggedright}
       }
      \toprule
     \textbf{Concepts}:  {\textbf{eye}, \textbf{hang}, \textbf{head}, \textbf{shut}, \textbf{squeeze}}  \\
      \textbf{Reference}:  A man \textbf{squeezes} his \textbf{eyes} \textbf{shut} and \textbf{hangs} his \textbf{head}.  \\

      \midrule
      \textbf{BART}: He \textbf{squeezes} her \textbf{head} \textbf{shut}, then grasps her \textbf{eyes} shut.      \\
      \midrule
      
    \textbf{Sentences retrieved by Ranker w/o distillation:} 
    (1) He kept his eyes closed tight. 
    (2) My eyes shut momentarily. \\
     \textbf{DKMR$^2$ + Ranker w/o distillation:} 
     He \textbf{squeezes} his \textbf{eyes} \
     \textbf{shut} and stares at the camera.   \\
     \midrule
     \textbf{Sentences retrieved by Distilled Ranker$_0$:} 
    (1) A baby with a blue shirt stretches while closing their eyes. 
    (2) a little baby closes his eyes as he has his head rubbed. \\
    \textbf{DKMR$^2$ + Distilled Ranker$_0$:} 
    A baby with a blue shirt \textbf{hangs} his \textbf{head} and \textbf{squeezes} his \textbf{eyes} \textbf{shut}.   \\
     \midrule
      
      \textbf{Sentences retrieved by Retriever$_0$:} 
        (1) Ca'daan shut his eyes tight. 
        (2) A young girl winks by closing her right eye lid. \\
     \textbf{DKMR$^2$ + Retriever$_0$:} 
        Someone \textbf{squeezes} her \textbf{head} \textbf{shut} and stares at him with a sad expression. \\
     \midrule
     \textbf{Sentences retrieved by Distilled Retriever$_1$:} 
    (1) happy laughing young casual woman with closed eyes holding the head. 
    \qquad \qquad(2) A baby in swaddling cloth is seen squeezing his eyes shut as he sneezes. \\
    \textbf{DKMR$^2$ + Distilled Retriever$_1$:} 
    A young woman with closed \textbf{eyes} holding the \textbf{head}. \\
     \bottomrule

    \end{tabular}
    \caption{The top two sentences retrieved by different retrieval models and sentences generated by our proposed model based on the retrieval sentences for the concepts extracted from the test set. 
    }\label{tab:example2}
\end{table*}

\subsection{Samples and Analysis}
As shown in Table \ref{tab:example2}, we can see that the sentence directly generated by BART violates our commonsense (e.g., `squeezes her head shut'), indicating the difficulty of this task. By comparison, DKMR$^2$+Ranker can generate a reasonable phrase (e.g., `squeezes his eyes shut'), mainly benefiting from the relevant information in the retrieval sentences (e.g., `keep his eyes closed' and `eyes shut'). We also observe that the sentences retrieved by the distilled ranker contain richer information (e.g., `A baby with a blue shirt' and `has his head rubbed'), making the generated sentence more realistic and informative. Similarly, we also see that sentences retrieved by the distilled retriever contain more information than those retrieved by the retriever without distillation. As a result, the sentence generated based on the retrieval sentences of the distilled retriever contains the missing concept `head', although it conveys `squeezes his eyes shut' with a synonymous expression `with closed eyes'. 

To summarize, both the distilled ranker and retriever can retrieve more relevant sentences compared with their counterparts without distillation, thus helping generate more reasonable sentences. 

\begin{table}[t] 
  \centering
 \footnotesize
   \begin{tabular}{
    m{0.13\textwidth}<{\raggedright}|
    m{0.08\textwidth}<{\centering}
    m{0.07\textwidth}<{\centering}
    m{0.095\textwidth}<{\centering}
    }
    \toprule
    \textbf{Models/Metrics}  &\textbf{BLEU-4} & \textbf{METEOR} & \textbf{ROUGE-L} \\
    \midrule
    BART   & 19.52 & 24.95 & 46.24  \\
    BART+TF-IDF   & 23.41 & 27.48  & 51.02 \\
    BART+DPR  & 23.74 & 27.49 & 51.48\\
    BART+Ranker &25.17 & 28.21 & 52.67 \\
    BART+Ranker$_0$ & \textbf{27.29} & \textbf{29.58} & \textbf{54.51} \\
    \bottomrule
 \end{tabular}
 \caption{Results of different models on the keyword generation test set.}\label{tab_keywords} 
\end{table} 

\subsection{Experiment on Keyword Generation} 
Following \citet{li-etal-2021-kfcnet-knowledge}, we further verify the proposed model on the keyword generation task. This task is meant to generate ads-relevant keywords matching the user intent. We collect some user inputs and the corresponding targeted keywords from an advertising platform. 
The training/dev/test set contains 5000/1000/1000 data instances. 
Below is a data instance selected from the training set: \\ 
source (user input): `good stock to invest now'  \\
target: `what are the best stocks to invest in right now'. 

We show the experiment results on the keyword generation test set in Table \ref{tab_keywords}. BART is the basic seq2seq baseline, where we directly fine-tune BART on the paired data. 
Similar to CommonGen, the retrieval-augmented methods first retrieve some relevant sentences from an external corpus (contains about 2M sentences), and then generate the target keywords based on the user input and the retrieved advertisements. 
The following four models are retrieval-augmented: 
BART+TF-IDF and BART+DPR denote BART using the data retrieved by TF-IDF and dense passage retriever to generate keywords. BART+Ranker means BART using the data retrieved by DPR and then re-ranked by the traditional ranker (i.e., KFCNet).  BART+Ranker$_0$ is our proposed method (i.e., DKMR$^2$ + Ranker$_0$), where the ranker is enhanced by the metric knowledge. From the above results, we can see our proposed method clearly outperforms previous retrieval-augmented baselines.

\section{Related Work}
\subsection{Text Retrieval.} Text retrieval aims to retrieve relevant texts for a given query. 
Traditional retrievers are implemented using sparse vector space models, such as TF-IDF and BM25, which have been used to retrieve the relevant passages for Open-domain question answering \cite{chen-etal-2017-reading}. 
Recently, with the success of large pre-trained models, such as BERT \cite{Devlin2019BERTPO} and RoBERTa \cite{Liu2019RoBERTaAR}, dense retrieval models \cite{karpukhin-etal-2020-dense, xiong2020approximate, qu-etal-2021-rocketqa} have surpassed the sparse vector space models, becoming the new de facto method. 
Dense passage retrievers are typically based on the dual-encoder architecture, which allows practitioners to compute the representation of each passage in the corpus and built indexes for them in advance. In this way, we only need to calculate the representation for the newly entered query and find the closest passage to the query, thus reducing the retrieval time. 

However, dual-encoder retrievers model the query and passage independently, thus failing to fully capture the fine-grained interactions between them. 
To solve this, BERT-based cross-encoder rankers \cite{wang-etal-2019-multi,DBLP:journals/corr/abs-1901-04085} are used to re-rank the retrieval passages of retrievers. 
Recently, the retrieve-then-rank pipeline has also been applied to solve CommonGen \cite{wang-etal-2021-retrieval-enhanced, li-etal-2021-kfcnet-knowledge, liu2021kgr}. 
Although rankers can effectively capture the relationships between the query and passage, the cross-encoder architecture makes it impractical to retrieve directly from the corpus. To alleviate this, recent work, such as AR2 \cite{AR2}, has focused on improving the retriever by distilling knowledge from the ranker. In this paper, we further extend this idea by distilling the order knowledge between the candidates and gold references to the ranker and retriever. 
\\
\subsection{Constrained Text Generation.} 
Constrained text generation is meant to generate text in a controlled way, 
such as generating text with the expected sentiment \citep{hucontrol}, style \citep{shen2017style,fu2018style}, length \cite{kikuchi2016controlling, fan2018controllable}, word definition \cite{he-yiu-2022-controllable}, or  
topic \cite{ficler2017controlling, Keskar2019CTRLAC}. 
Lexically constrained text generation is another kind of controllable text generation, aiming to incorporate some specific keywords into outputs. 
Researchers solve this task by controlling the decoding process \cite{ Hokamp2017LexicallyCD,post-vilar-2018-fast} or refining candidate outputs iteratively \cite{sha-2020-gradient,he2021xlentmcmc, he2021parallel}. CommonGen is related to lexically text generation. The main differences between them are twofold: (1) CommonGen does not force the given keywords/concepts to appear in outputs; (2) CommonGen proposes challenges to the compositional generalization ability of models. 
\\

\section{Conclusions}
This work presents DKMR$^2$, a novel retrieval-augmented model for generative commonsense reasoning. 
Unlike previous work, DKMR$^2$ enhances the retrieval module with the guidance of the evaluation metric. 
To be concrete, DKMR$^2$ first distills the order knowledge from the metric to the ranker and then teaches the key points summarized by the distilled ranker to the retriever. 
As a result, DKMR$^2$ achieves the state-of-the-art results on CommonGen. More importantly, DKMR$^2$ narrows the performance gap between the ranker and retriever, resulting in DKMR$^2$ with a distilled retriever being  better than the previous baselines. 



\section*{Limitations}
This work mainly focuses on improving the retrieval module with metric-guided distillation. 
There may be two possible limitations in our study. 
The first concerns the model size. Given the cost of  retrieving, our retrieval module is based on the base model of BERT. Applying our proposed method to larger models, such as BERT-large and RoBERTa-large, may lead to further improvement.  
The second limitation is that we verify the effectiveness of this method only on the generative commonsense reasoning task. However, our proposed method can also be extended to other knowledge-intensive generation tasks, such as open-domain question answering and fact verification. 
In the future, we plan to use the metric-guided distillation to improve the retrieval modules of these tasks.

\section*{Acknowledgements}
This project is supported by the funding from HKUSCF FinTech Academy. 
We would like to thank the anonymous reviewers for their constructive and informative feedback on this work.

\bibliography{emnlp2022}
\bibliographystyle{acl_natbib}

\appendix

\clearpage
\section{Retriever and Ranker}\label{framework}
\begin{figure}
  \centering
    \subfigure[Retriever]{
      \centering
      \includegraphics[width=0.5\textwidth]{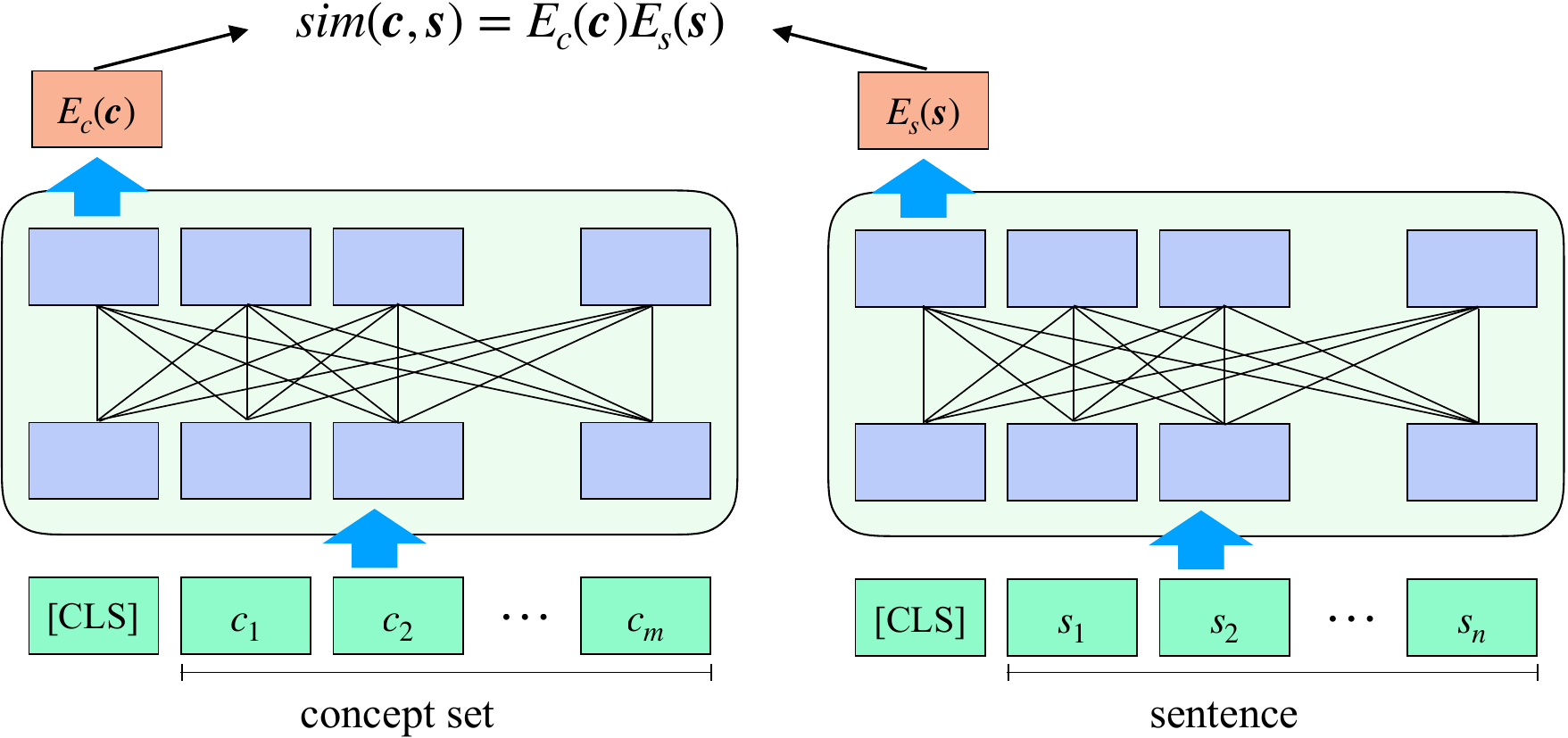} 
    }
    \subfigure[Ranker]{
      \centering
      \includegraphics[width=0.46\textwidth]{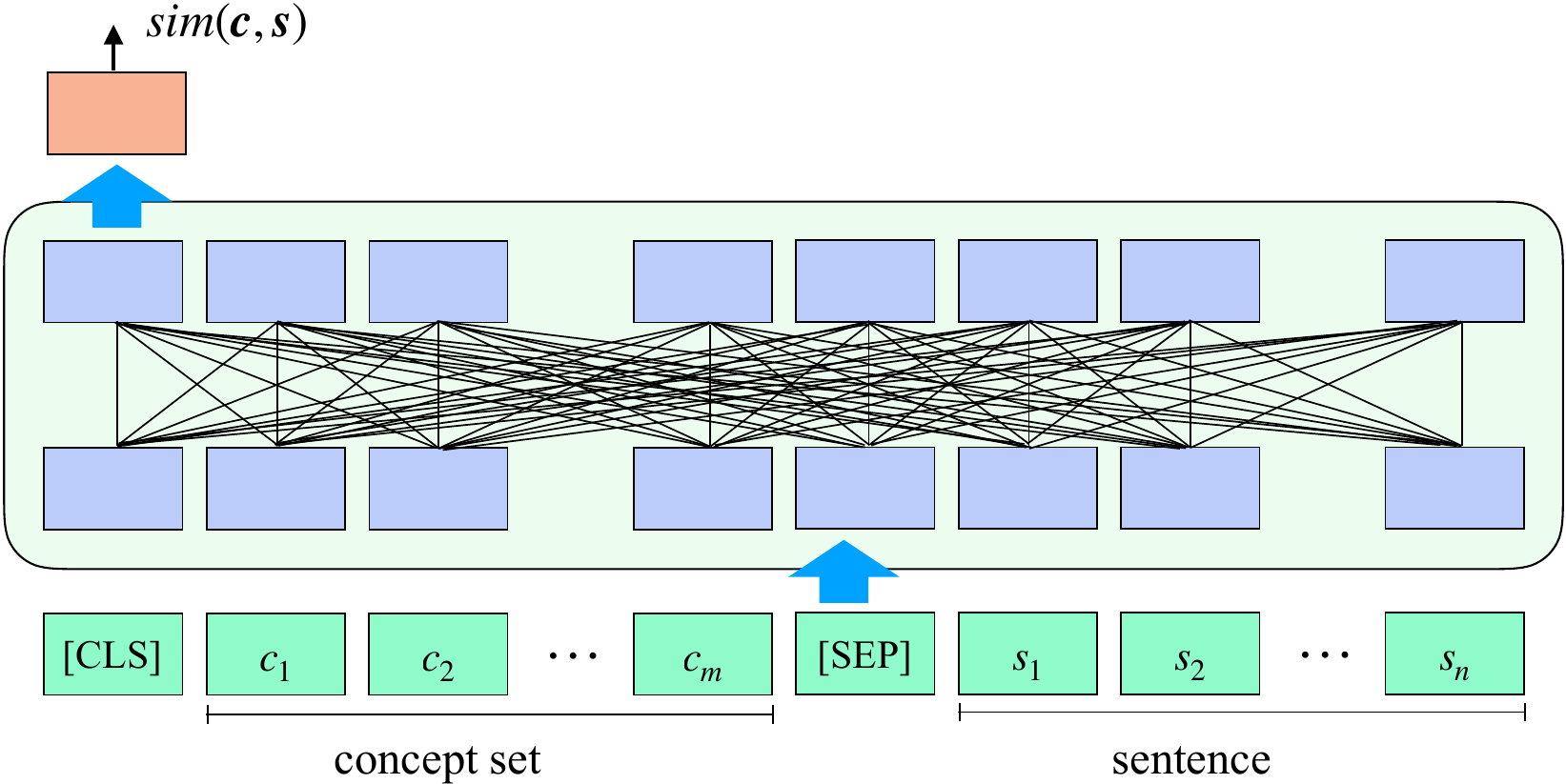} 
    }
      \caption{ 
            Illustration of the retriever and ranker used in our model. (a) For the dual-encoder retriever, the query and sentence are encoded independently by two BERT-based models. (b) For the cross-encoder ranker,  the concatenated query and sentence are jointly encoded by a BERT-based model. 
      }
      \label{retriever_ranker}
\end{figure}

We show the architecture of the retriever and ranker in Figure \ref{retriever_ranker}. 

\section{Implementation Details}\label{imp}
\subsection{Retriever$_0$}
The dual-encoder retriever is initialized from the bert-base-cased model. 
During the warm-up stage, we optimize the retriever using the Adam optimizer \citep{kingma2015adam}  with an initial learning rate of $2e-5$ and batch size of 1000. At this stage, each concept set input in the batch is paired with one positive sentence, in-batch negatives, and one hard negative sentence (For each concept set, we use concept matching to  prepare 100 sentences containing the most concepts as the hard negative pool in advance. During training, we randomly sample one sentence from the pool as the hard negative). We evaluate the model on the validation every epoch and select the checkpoint with the highest Recall of top-1, i.e., R@1 (We expect to evaluate whether the model can select the positive sentence from the candidate sentences). As for the retrieval period, we use the top $K=100$ sentences returned by the retriever as candidates. 

\subsection{Ranker$_0$}
Similar to the retriever, the ranker is also initialized from the bert-base-cased model. During training, we set the batch size to 80, and the learning rate to $2e-5$. 
When training, each concept set is paired with one positive sentence, 10 hard negatives from the   top $K=100$ candidate sentences returned by the warm-up retriever (i.e., $N_1=11$ in Equation \ref{eq3}). 
We evaluate the ranker every 100 steps and choose the checkpoint with the highest R@1 on the validation set. Unlike the warm-up stage of the retriever, we only use the positive sentence during training but discard it during evaluation. In other words, during evaluation, the candidate pool does not contain the positive one and the model is expected to choose the negative sentence (i.e., $N_1=10$) that most resembles the reference sentences of the given concepts.

\subsection{Retriever$_1$}
During the distillation stage, we continue to fine-tune the retriever with the guidance of the ranker. At this stage, each concept set is paired with one positive sentence, 10 hard negatives from the same candidate pool with the warm-up stage (i.e., $N_2$=11 in Equation \ref{eq4}), but without in-batch negatives. During training, the learning rate is set to $2e-5$, with a batch size of 200. As for evaluation, we resort to the same settings with the ranker. We use BLEU as the distillation metric $M$ for the ranker and retriever. 

\subsection{Generation Model}
We initialize the generation model with BART-base. We feed the top $k=2$ retrieved sentences for each concept set to BART to help generate target sentences. $k$ is searched from $\{1, 2, 3, 4, 5\}$. 
During training, we fine-tune the generation model with an initial learning rate of 3e-5 for up to 20 epochs, and set the batch size to 400. We evaluate the model every epoch and choose the checkpoint with the lowest negative log-likelihood (NLL) loss on the validation set. 
During inference, we run beam search decoding with beam width = 5 and max decoding length = 60 on the generation model.

We use the Adam optimizer with slightly different learning rates for all models, where the learning rate is searched from $\{5e-6, 1e-5, 2e-5, 3e-5, 4e-5, 5e-5\}$. 
We also use early stopping with the patience of two and choose the checkpoints based on their performance on the validation set during training. 
We implement all models with the HuggingFace Transformers library \citep{Wolf2019HuggingFacesTS} and  conduct all experiments on 4 NVIDIA Tesla V100 GPUs with 32 GB memory.

\section{Effect of Hard Negatives on Retriever$_0$}\label{hardneg}
As stated in Section \ref{retriever}, when warming up Retriever$_0$, we consider two methods to create hard negative sentences. To test their effect on the generation model, we train Retriever$_0$ with each method and then create candidate sentence pool $P_0$ with the warm-up retriever. Finally, the generation model generates target sentences based on $P_0$. As shown in Table \ref{tab_result6}, the generation model performs slightly better when using concept  matching to create hard negatives. Therefore, we resort to concept matching to create hard negatives for Retriever$_0$.

\begin{table}[t] 
  \centering
 \footnotesize
   \begin{tabular}{
    m{0.18\textwidth}<{\raggedright}|
    m{0.08\textwidth}<{\centering}
    m{0.05\textwidth}<{\centering}
    m{0.05\textwidth}<{\centering}
    }
    \toprule
    \textbf{Hard Negatives}  &\textbf{BLEU-4} & \textbf{CIDEr} & \textbf{SPICE} \\
    \midrule
    Concept Matching  & \textbf{55.50} & \textbf{21.96} & 
    \textbf{39.93} \\
    TF-IDF & 54.38 & 21.51 & 39.89 \\
    \bottomrule
 \end{tabular}
 \caption{ Results of training Retriever$_0$ with different methods to create hard negative sentence on the CommonGen test set (v1.0). 
 }\label{tab_result6} 
\end{table}

\end{document}